\documentclass[runningheads]{llncs}
\usepackage[T1]{fontenc}
\usepackage{graphicx}
\usepackage{amsmath}
\usepackage{subcaption}
\usepackage{xcolor}
\usepackage[
maxnames=5,                         
sorting=none
]{biblatex}
\usepackage{booktabs}    
\usepackage{caption}     
\usepackage{graphicx}    
\usepackage{xcolor}
\usepackage{pict2e}

\newsavebox{\ORCIDlogo}
\savebox{\ORCIDlogo}{%
\setlength{\unitlength}{\dimexpr 1em/256\relax}%
\begin{picture}(256,256)%
  \color[HTML]{A6CE39}\put(128,128){\circle*{256}}%
  \color{white}%
  \put(78.6,199.2){\circle*{20}}%
  \moveto(70.9,176,9)\lineto(86.3,176,9)\lineto(86.3,69.8)\lineto(70.9,69.8)%
  \closepath\fillpath%
  \moveto(108.9,176.9)\lineto(150.5,176.9)%
  \curveto(190.1,176.9)(207.5,148.6)(207.5 ,123.3)%
  \curveto(207.5,95,8)(186,69.7)(150.7,69.7)%
  \lineto(108.9,69.7)%
  \closepath\fillpath%
  \color[HTML]{A6CE39}%
  \moveto(124.3,83.6)\lineto(148.8,83.6)%
  \curveto(183.7,83.6)(191.7,110.1)(191.7,123.3)%
  \curveto(191.7,144.8)(178,163)(148,163)%
  \lineto(124.3,163)%
  \closepath\fillpath%
\end{picture}%
}

\newcommand\orcidicon[1]{\href{https://orcid.org/#1}{\usebox{\ORCIDlogo}}}

\usepackage{hyperref} 

\begin{document}
\title{Towards Label-Free Brain Tumor Segmentation: Unsupervised Learning with Multimodal MRI}
\titlerunning{Towards Label-Free Brain Tumor Segmentation}
%
\author{Gerard Comas-Quiles\inst{1} \and
Carles Garcia-Cabrera\inst{2,4}\orcidicon{0000-0001-8139-9647} \and
Julia Dietlmeier\inst{3, 4}\orcidicon{0000-0001-9980-0910} \and \newline Noel E. O'Connor\inst{3, 4}\orcidicon{0000-0002-4033-9135} \and Ferran Marques \inst{1}\orcidicon{0000-0001-8311-1168}}
\authorrunning{G. Comas et al.}
\institute{Universitat Politècnica de Catalunya (UPC), Barcelona, Spain
\and University College Dublin (UCD), Dublin, Ireland
\and Dublin City University (DCU), Dublin, Ireland
\and Insight Research Ireland Center For Data Analytics, Dublin, Ireland
}

\maketitle              

\begin{abstract}
Unsupervised anomaly detection (UAD) presents a complementary alternative to supervised learning for brain tumor segmentation in magnetic resonance imaging (MRI), particularly when annotated datasets are limited, costly, or inconsistent. In this work, we propose a novel Multimodal Vision Transformer Autoencoder (MViT-AE) trained exclusively on healthy brain MRIs to detect and localize tumors via reconstruction-based error maps. This unsupervised paradigm enables segmentation without reliance on manual labels, addressing a key scalability bottleneck in neuroimaging workflows. Our method is evaluated in the BraTS-GoAT 2025 Lighthouse dataset, which includes various types of tumors such as gliomas, meningiomas, and pediatric brain tumors. To enhance performance, we introduce a multimodal early-late fusion strategy that leverages complementary information across multiple MRI sequences, and a post-processing pipeline that integrates the Segment Anything Model (SAM) to refine predicted tumor contours. Despite the known challenges of UAD, particularly in detecting small or non-enhancing lesions, our method achieves clinically meaningful tumor localization, with lesion wise Dice Similarity Coefficient of 0.422 (Whole Tumor), 0.279 (Tumor Core), and 0.289 (Enhancing Tumor), and an anomaly Detection Rate of 89.4\%. These findings highlight the potential of transformer-based unsupervised models to serve as scalable, label-efficient tools for neuro-oncological imaging. The code for this project will be publicly available on \href{https://github.com/gerardco7/Towards-Label-Free-Brain-Tumor-Segmentation}{GitHub}.

\end{abstract}

\section{Introduction}
Advancements in medical imaging, particularly magnetic resonance imaging (MRI), have significantly improved the early detection and characterization of brain tumors. However, interpreting MRI scans remains a labor-intensive and error-prone process. Studies report that 5--10\% of neuroimaging interpretations may overlook critical abnormalities \cite{doi:10.2214/AJR.13.11493, doi:10.1148/rg.2015150023}, underscoring the need for robust and automated diagnostic tools that can support human interpretation.

The Brain Tumor Segmentation (BraTS) Challenge \cite{menze2014multimodal,bakas2017advancing} has played a central role in advancing brain tumor segmentation techniques, particularly for gliomas. Since its inception in 2012, BraTS has progressively expanded to include additional tumor types and imaging modalities, and to increase clinical relevance. Supervised deep learning methods—especially U-Net architectures \cite{ronneberger2015u}, 3D CNNs \cite{isensee2021nnu}, and more recently, transformer-based models \cite{hatamizadeh2022unetr,chen2021transunet}—have achieved state-of-the-art performance when trained on large, annotated datasets.

Despite these advancements, reliance on manual annotations poses significant limitations in real-world clinical settings. Annotation is not only time-consuming and costly, but also susceptible to inter-observer variability \cite{article_de_sutter}, particularly across institutions. These challenges become even more pronounced when dealing with a small number of lesions \cite{VANDERLOO2025111874}, such as rare tumor types or low-resource environments.

To move beyond models limited to detecting only specific tumor types, the BraTS challenge encouraged research into generalization methods \cite{brats2023_crossmoda}. However, no submissions adopted fully unsupervised learning, underscoring both the technical challenges and the unexplored potential of unsupervised anomaly detection (UAD) in brain tumor segmentation.

In this work, we propose a fully unsupervised brain tumor segmentation pipeline designed for the BraTS Generalizability Across Tumors (GoAT) 2025 challenge. Our method combines a multimodal Vision Transformer Autoencoder (MViT-AE) with a novel postprocessing stage based on the Segment Anything Model (SAM) \cite{kirillov2023segment}. The model is trained solely on healthy brain MRIs to learn priors of normal anatomy. During inference, reconstruction errors reveal anomalous regions such as tumors, which are then refined into segmentation masks using morphological operations and SAM-based region proposals.

\begin{figure}[htbp]
    \centering
    \includegraphics[width=1\textwidth]{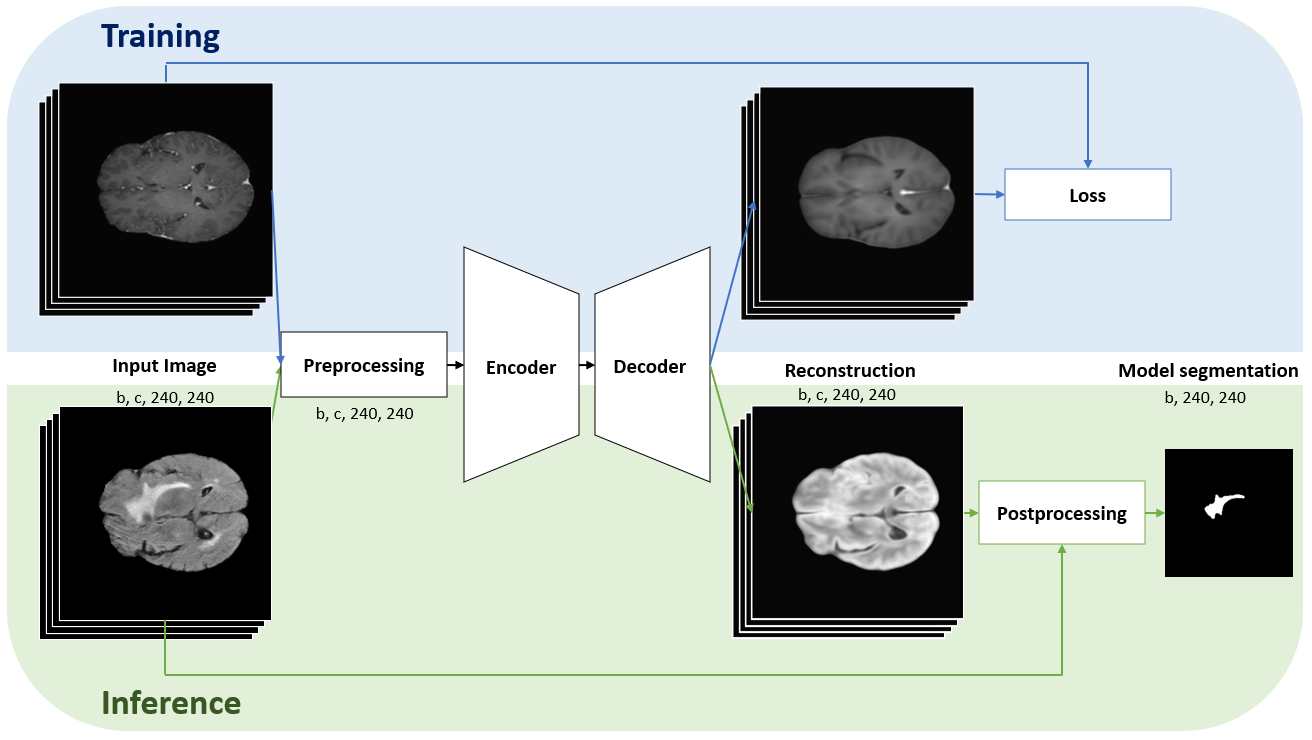}
    \caption{Overview of the proposed Unsupervised Anomaly Segmentation pipeline, comprising preprocessing, transformer-based reconstruction, and SAM-guided postprocessing. Multimodal MRI slices are used to learn healthy anatomical priors, and reconstruction deviations are used to infer tumor regions.}
    \label{fig:methodology_pipeline}
\end{figure}

An overview of the proposed pipeline is shown in Figure~\ref{fig:methodology_pipeline}, consisting of three stages: multimodal fusion and preprocessing, transformer-based image reconstruction, and SAM-guided postprocessing.

\subsection{Related Work}
UAD has emerged as a promising strategy for medical image analysis in data-scarce settings. Traditional approaches rely on convolutional autoencoders trained to reconstruct healthy anatomical structures, flagging abnormalities via high reconstruction error \cite{baur2021autoencoders}. However, these models often suffer from blurry outputs and poor localization.

Variational autoencoders (VAEs) introduced probabilistic modeling to capture latent uncertainty \cite{zimmerer2019context}, but their reconstructions lack structural detail, limiting segmentation accuracy. Extensions using perceptual losses, adversarial training \cite{chen2018unsupervised}, or memory modules \cite{venkataramanan2020attention} have sought to improve anomaly saliency and boundary sharpness.

More recent UAD research has focused on learning disentangled representations through contrastive techniques \cite{10677936}, enabling separation of normal and abnormal features in latent space. These approaches often leverage structural similarity (SSIM) or learned feature distances as anomaly scores.

Despite these efforts, most prior work uses CNN backbones with limited receptive fields, hindering global context modeling. Transformer-based architectures, such as UNETR \cite{hatamizadeh2022unetr} and TransUNet \cite{chen2021transunet}, have demonstrated success in supervised segmentation, but remain largely unexplored in the unsupervised setting.

\subsection{Contributions}
This paper presents the first fully unsupervised segmentation pipeline evaluated in the BraTS-GoAT setting. Our main contributions include:

\begin{itemize}
    \item \textbf{Multimodal Vision Transformer Autoencoder (MViT-AE):} We introduce a transformer-based autoencoder capable of learning global contextual representations from fused multimodal MRI slices. This addresses the spatial limitations of prior convolutional UAD models and improves reconstruction fidelity.
    
    \item \textbf{Novel early–late fusion strategy:} We stack all four MRI modalities as input channels and later, after postprocessing, fuse T1c with T2f to exploit their complementary information, leading to more consistent and robust feature representations.
    
    \item \textbf{SAM-based postprocessing:} We incorporate SAM into the anomaly detection pipeline to transform coarse reconstruction error maps into high-quality segmentation masks. While semi-supervised SAM extensions (e.g., SemiSAM for heart MRI \cite{zhang2024semisamenhancingsemisupervisedmedical}) have been explored, to the best of our knowledge, this is the first application of SAM in an unsupervised medical segmentation setting.
\end{itemize}

\section{Methods}
All experiments were conducted on a workstation equipped with an Intel Core i9-9900K CPU @ 3.80 GHz, 64 GB of RAM, and a NVIDIA GTX 2080 Ti GPU, using the PyTorch framework.

\subsection{Data}
We used the BraTS-GoAT 2025 dataset, which integrates multiple BraTS challenges to enable evaluation of cross-tumor generalization. The training set includes adult gliomas~\cite{articleBratsGli,articleBratsGli2,Bakas2017_GBM,Bakas2017_LGG,baid2021rsnaasnrmiccaibrats2021benchmark}, meningiomas~\cite{labella2023asnrmiccaibraintumorsegmentation}, and brain metastases~\cite{moawad2024braintumorsegmentationbratsmets}. The validation set extends this with additional cohorts, namely gliomas from Sub-Saharan Africa~\cite{adewole2023braintumorsegmentationbrats} and pediatric brain tumors~\cite{kazerooni2024braintumorsegmentationbrats}. The test set composition is not disclosed by the organizers to ensure unbiased evaluation. Each case consists of four mpMRI modalities (T1c, T1n, T2f, T2w) with expert annotations in NIfTI format. The training set consists of 1,352 multiparametric MRI volumes (mpMRI), each containing four modalities (T1c, T1n, T2f, and T2w), with corresponding expert annotations in NIfTI format. Since unsupervised learning requires training on images without anomalies, each volume was split into two pseudo-volumes: one containing slices with anomalies (used for internal validation) and the other containing only healthy brain slices (used for training). This resulted in a training set of 98,354 2D images, each of size $240 \times 240$ pixels.

Preprocessing is critical for consistent model learning. MRI intensity values vary widely due to acquisition conditions and lack an absolute physical meaning. We applied z-score normalization on a per-volume basis to standardize intensities while preserving the original distribution, ensuring anomalies remain visible during training.

\subsection{Model}
Our model, MViT-AE, is based on a 2D Vision Transformer Autoencoder originally designed for unsupervised anomaly detection in industrial images \cite{Mishra_2021}. We retain the ViT encoder, adapting it to handle mpMRI inputs, and use a custom decoder consisting of 6 convolutional layers. 

\begin{figure}[htbp]
    \centering
    \includegraphics[width=1\textwidth]{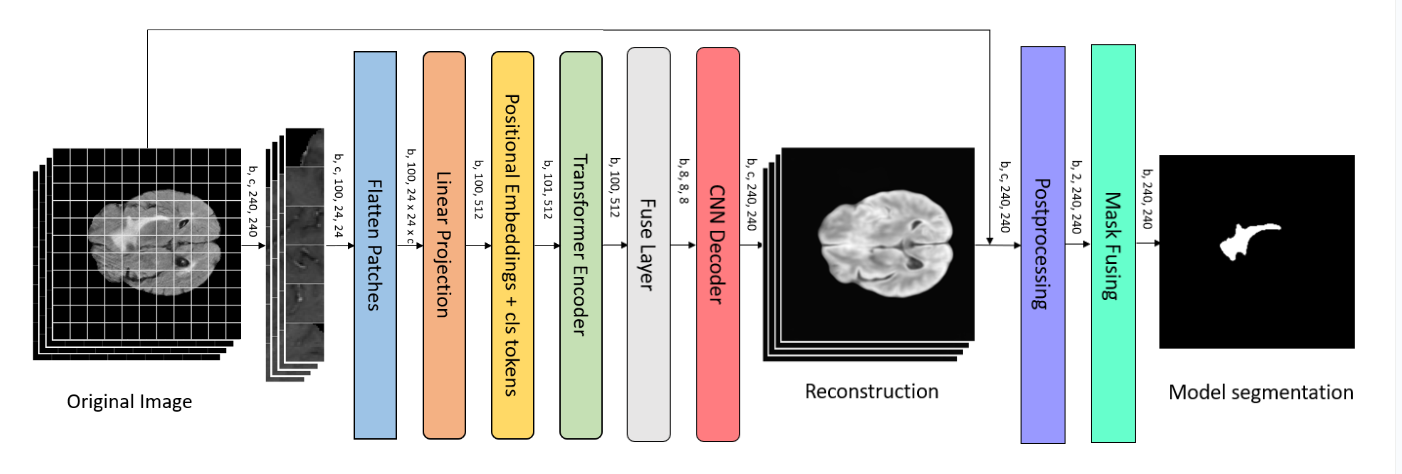}
      \caption[Multimodal ViT-AE architecture]{Overview of the Multimodal ViT-AE (MViT-AE) architecture. The encoder processes multiple input modalities through transformer layers, followed by a fusion layer. The decoder reconstructs each modality, and postprocessing with fusion masking is used to generate the final segmentation.}
    \label{fig:MViT}
\end{figure}

As shown in Figure~\ref{fig:MViT}, each modality is split into $24 \times 24$ patches, which are concatenated along the channel dimension, resulting in input vectors of size $24 \times 24 \times 4$. Each patch is flattened into a 576-dimensional vector and projected into a 512-dimensional embedding. Positional embeddings preserve spatial context. The sequence of patch embeddings passes through a Transformer encoder with 6 layers, each having 8 attention heads and a feed-forward network with 1024 hidden units. The encoder output is compressed via a fully connected fusion layer (MLP) into a 512-dimensional latent vector. The decoder reshapes this vector into an $(8,8,8)$ feature map and progressively upsamples spatial dimensions while reducing channels to reconstruct all four modalities, producing outputs that match the input size. The model contains approximately 40.7 million parameters.

\subsection{Training}

We trained the model using the Adam optimizer with a fixed learning rate of 0.0001 and a weight decay of 0.0001, batch size of 32, for 100 epochs. We did not use a learning rate scheduler, data augmentation, or early stopping. The model checkpoint was saved only when the validation loss improved to keep the best-performing version. Future work could focus on hyperparameter tuning to enhance results.

The loss function combines Mean Squared Error (MSE) and Structural Similarity Index Measure (SSIM) losses:
\[
L = L_{Rec} + \alpha \, L_{SSIM},
\]
where \(L_{SSIM} = 1 - \text{SSIM}(x, \hat{x})\), and \(\alpha = 100\) was chosen empirically to balance the contributions of both losses effectively.

Gaussian noise (mean=0, std=0.2) was added during training to improve robustness and encourage the model to learn stable features by simulating subtle variations.

\subsection{Postprocessing}
Since the model produces reconstructed images rather than direct segmentations, postprocessing is essential for accurate tumor delineation.

We compute residual maps by subtracting the reconstructed images from the originals, using the signed difference to focus on hyperintense, poorly reconstructed areas, which reduces noise compared to absolute differences~\cite{baur2021autoencoders}.

\begin{figure}[htbp]
    \centering
    \includegraphics[width=1\textwidth]{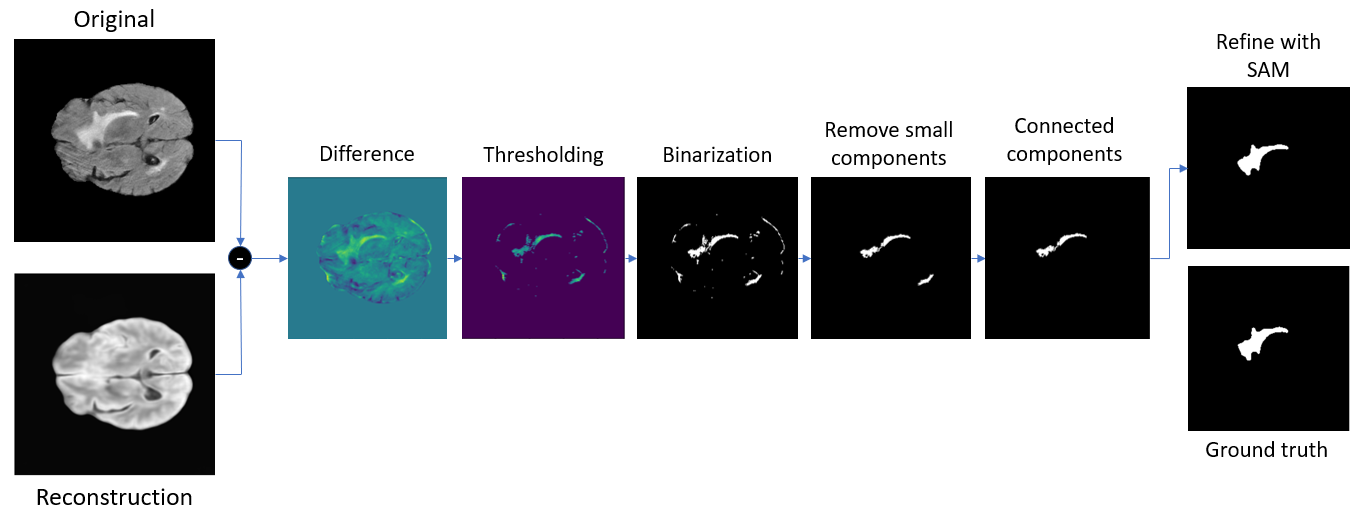}
      \caption[Postprocessing Pipeline]{Postprocessing pipeline: The original and reconstructed images are subtracted to highlight differences. After thresholding and binarization, small irrelevant regions are removed. A 3D connected component algorithm selects the largest region, which is then refined using the Segment Anything Model (SAM) for improved segmentation accuracy.}
    \label{fig:postprocessing}
\end{figure}

The postprocessing steps (Figure~\ref{fig:postprocessing}) include:

\begin{enumerate}
    \item \textbf{Thresholding}: We set the threshold to the maximum between 20\% of the highest residual value and a fixed value of 1.2. This balances adaptability and robustness, and was empirically chosen to optimize validation metrics.
    \item \textbf{Binarization}: Using Otsu's automatic thresholding method \cite{otsu1979threshold}, we convert the thresholded residual map into a binary segmentation mask, separating tumor from non-tumor regions.
    \item \textbf{Removal of small objects}: To reduce false positives caused by small, isolated noise regions, we apply a morphological opening and closing with structure element of size 1.
    \item \textbf{3D Connected components}: Because tumors typically form contiguous structures in 3D, we identify connected components in the full volume and keep only the largest one, which is most likely the true tumor, discarding smaller irrelevant regions.
    \item \textbf{Refinement with SAM (Segment Anything Model)}: To further improve segmentation quality, we refine the initial segmentation mask using SAM, a powerful general-purpose segmentation model. SAM is capable of producing high-quality object masks from various prompts such as points, bounding boxes, or coarse masks.
\end{enumerate}

\begin{figure}[htbp]
    \centering
    \begin{subfigure}[b]{0.45\textwidth}
        \centering
        \includegraphics[width=\textwidth]{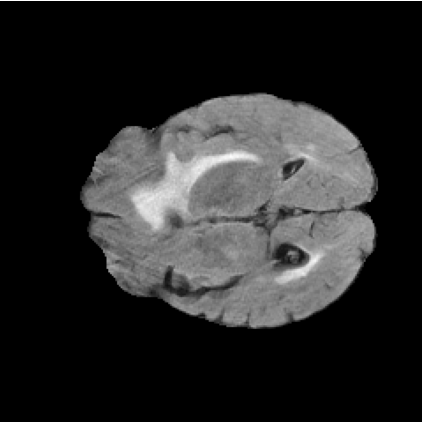}
        \caption{Original MRI slice}
        \label{fig:sam_input}
    \end{subfigure}
    \hfill
    \begin{subfigure}[b]{0.45\textwidth}
        \centering
        \includegraphics[width=\textwidth]{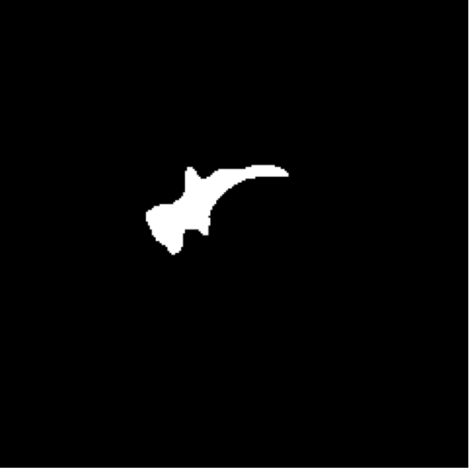}
        \caption{Ground truth}
        \label{fig:sam_initial_mask}
    \end{subfigure}

    \vskip\baselineskip

    \begin{subfigure}[b]{0.45\textwidth}
        \centering
        \includegraphics[width=\textwidth]{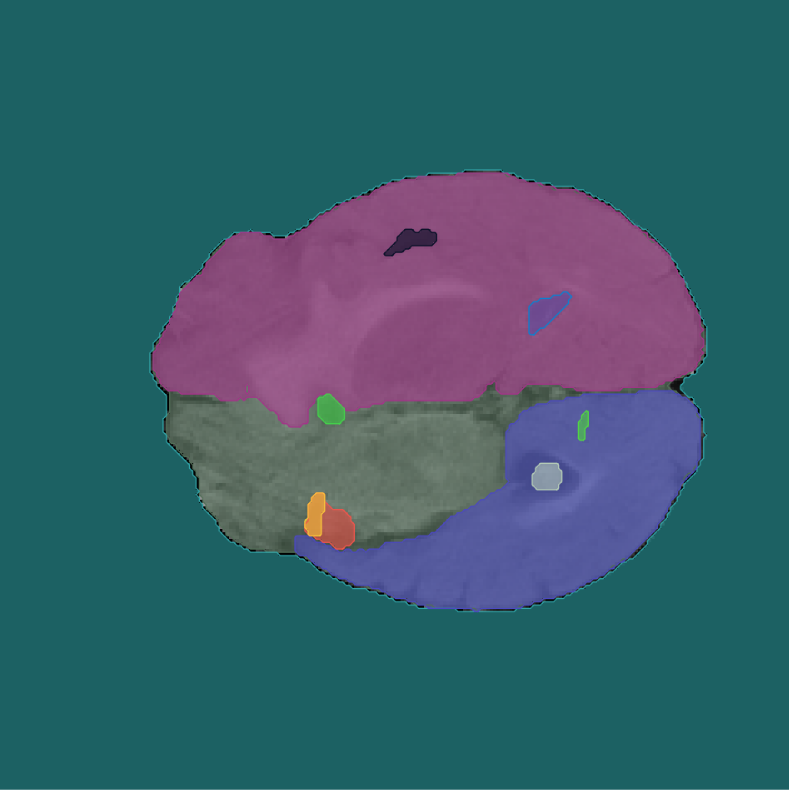}
        \caption{Initial SAM segmentation}
        \label{fig:sam_mask}
    \end{subfigure}
    \hfill
    \begin{subfigure}[b]{0.45\textwidth}
        \centering
        \includegraphics[width=\textwidth]{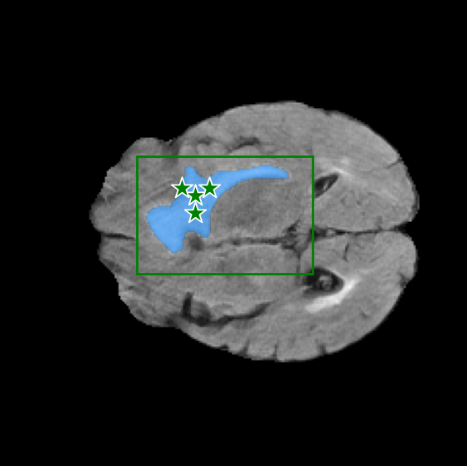}
        \caption{Prompt SAM segmentation mask}
        \label{fig:sam_refined_mask}
    \end{subfigure}

    \caption{
    Comparison of segmentation stages using the Segment Anything Model (SAM). 
    (a) Original MRI slice. 
    (b) Ground truth mask showing the expert-annotated tumor. 
    (c) Initial SAM segmentation without prompts, showing poor accuracy. 
    (d) Refined SAM segmentation using bounding box and point prompts, with clearer tumor boundaries.
    }
    \label{fig:sam_subplot}
\end{figure}

As shown in Figure~\ref{fig:sam_subplot}, applying SAM directly to MRI slices without any guidance results in segmentations that lack medical relevance and fail to accurately capture tumor regions. To address this, we use the initial segmentation mask to generate prompts for SAM:
 \begin{itemize}
        \item We compute a tight bounding box enclosing the predicted tumor region.
        \item We randomly sample five foreground points from within this region.
        \item These prompts, consisting of a bounding box and foreground points, are provided to SAM to guide the segmentation.
    \end{itemize}

We accept SAM's refined mask only if it achieves a confidence score above 90\%; otherwise, we retry with new points up to three times before falling back to the original mask. This approach leverages SAM's strengths to produce smoother, more accurate boundaries and reduce noise artifacts.

\subsection{Mask Fusing}
Each MRI modality produces its own reconstruction and residual map. After post-processing each modality independently, T1c-based segmentation identifies the Enhancing Tumor (ET), while the difference between T2f and T1c segmentations outlines the Surrounding Non-Enhancing FLAIR Hyperintensity (SNFH) region. Any holes or gaps within the tumor area—regions not classified as either ET or SNFH—are assumed to be Non-Enhancing Tumor (NET). This approach allows us to segment all three main tumor subregions without any manual labels, demonstrating the power of multimodal fusion. To our knowledge, no previous unsupervised method has achieved this level of subregion segmentation.

\section{Results}

\subsection{Quantitative Results}
The following quantitative results are based on the test set from the BraTS-GoAT 2025 challenge, which includes 451 mpMRI volumes. We processed the complete volumes without cropping or patching, but using the same preprocessing as the training dataset.

We evaluated segmentation performance using a lesion wise Dice Similarity Coefficient (DSC), focusing on five BraTS tumor subregions: ET, NET, SNFH, Tumor Core (ET + NET = TC), and Whole Tumor (TC + SNFH = WT). In addition to DSC, we report the Detection Rate (DR), which measures how often the model detects any part of the tumor. Specifically, DR is computed as the number of cases where the DSC for the WT is greater than zero, divided by the total number of test cases.

Table~\ref{tab:sam_comparison} shows the metrics for our model, MViT-AE, both with and without SAM refinement.
\begin{table}[htbp]
    \centering
    \caption{Lesion wise DSC and DR of MViT-AE With and Without SAM Refinement}
    \label{tab:sam_comparison}
        \begin{tabular}{@{}lcccccc@{}}
            \toprule
            \textbf{Method} & \textbf{DSC ET} & \textbf{DSC NET} & \textbf{DSC SNFH} & \textbf{DSC TC} & \textbf{DSC WT} & \textbf{DR \%}\\
            \midrule
            MViT-AE & 0.289 & 0.088 & 0.473 & 0.279 & 0.422 & 89.4\\
            MViT-AE + SAM & 0.204 & 0.082 & 0.524 & 0.231 & 0.371 & 89.1\\
            \bottomrule
        \end{tabular}
\end{table}

Overall, the original MViT-AE model without SAM performed better for most DSC scores. The only exception was the SNFH, where the SAM-refined model scored higher, improving from 0.473 to 0.524.

Regarding DR, both models show very similar results: 89.4\% for MViT-AE and 89.1\% for MViT-AE + SAM. This similarity is expected, as the SAM module refines the segmentation mask generated by MViT-AE, but does not introduce new detections i.e. it improves the shape or boundaries of existing masks, but cannot recover tumors missed by the initial model.

These results suggest that while SAM may enhance the segmentation quality in certain regions (e.g., SNFH), it does not consistently improve overall performance and may even degrade it in more localized tumor subregions.

\subsection{Qualitative Results}
In this section, we present qualitative examples to better understand how the model leverages the different MRI modalities for tumor detection, and how the SAM optimization helps refine the segmentation.

\begin{figure}[htbp]
    \centering
    \includegraphics[width=1\textwidth]{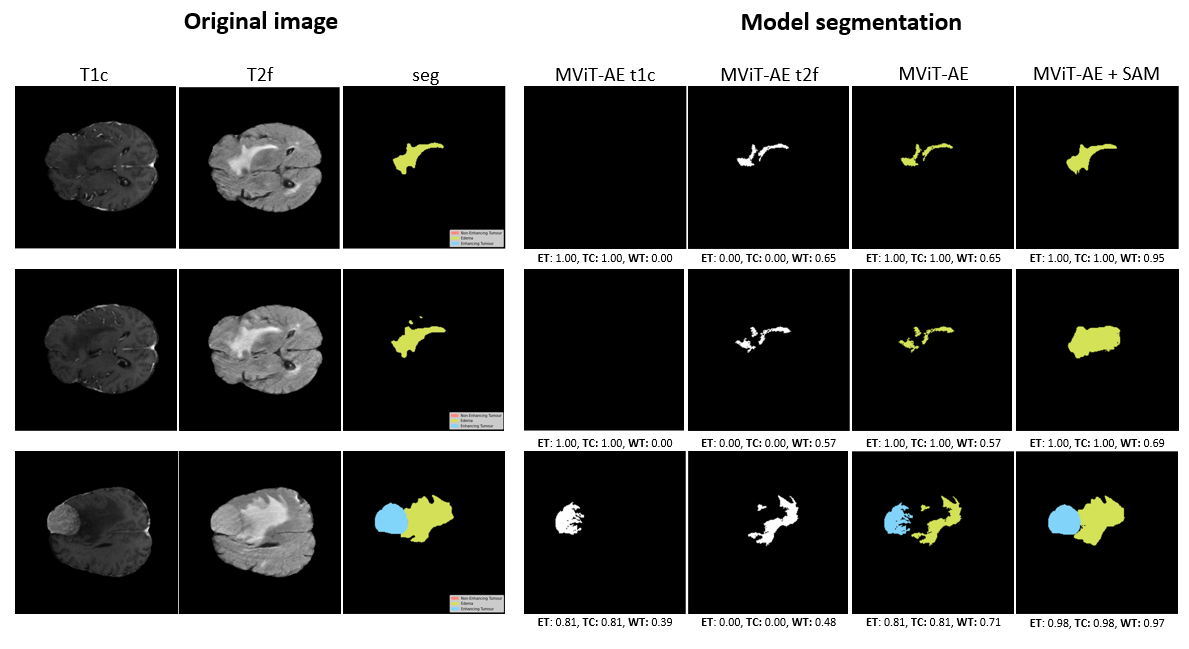}
    \caption[Qualitative segmentation results for Volume 18.]{Qualitative segmentation results for volume \texttt{00018} from the BraTS-MEN 2023 dataset. The first three columns show the original input modalities (T1c, T2f) and the ground truth. The following four columns present the outputs of different model branches: MViT-AE with only T1c, only T2f, both combined, and the combination refined using SAM. Each row corresponds to a different slice from the same volume.}
    \label{fig:qualitative_results}
\end{figure}

The results in Figure~\ref{fig:qualitative_results} highlight the complementary nature of the modalities. In the column corresponding to the T1c-only model, ET (\textcolor{blue}{\textbf{blue}}) core is clearly visible, but SNFH (\textcolor{green}{\textbf{green}}) is largely missed. In contrast, the T2f-only model captures SNFH region well, but is less effective at identifying the enhancing part of the tumor. When both modalities are combined, the resulting segmentation benefits from the strengths of each. This fusion results in a more complete and accurate tumor representation. 

Finally, the last column shows the results after applying SAM for post-processing. We observe that the segmentation becomes more anatomically consistent, with smoother boundaries and less pixel-level noise. This refinement significantly improves the visual quality and realism of the predicted masks. However, SAM can occasionally introduce false positives, as seen in the second row, where a part of the brain is mistakenly segmented as tumor. \\

\section{Discussion and Conclusion}
We presented an unsupervised brain tumor segmentation pipeline based on MViT-AE, trained exclusively on healthy brain MRI scans. By leveraging reconstruction errors and refining the outputs with SAM, the method localizes tumors without requiring manual annotations.

Our best model achieved a Detection Rate of 89.4\%. Most missed cases involved hypointense tumors in T1c, suggesting a need for improved sensitivity to such patterns. While performance remains below state-of-the-art supervised methods, results highlight the promise of unsupervised anomaly detection for this task.

A key limitation is the postprocessing step, which retains only the largest 3D component. This restricts outputs to a single segmentation per volume and negatively impacts lesion-wise metrics when multiple anomalies are present. Another limitation arises from the use of SAM: it refines accurate masks by capturing fine details (e.g., in SNFH regions) but can amplify errors when the initial mask is poor (e.g., in ET regions).

Despite these challenges, the approach reduces reliance on costly labeled datasets and shows potential to generalize across tumor types and imaging protocols. Clinically, it could assist radiologists by flagging suspicious regions, especially in data-scarce settings. 

Future work will focus on increasing sensitivity to subtle anomalies, integrating additional imaging sequences, improving computational efficiency, and exploring semi-supervised fine-tuning. Further validation on healthy control datasets and more interpretable outputs will also be essential for clinical translation.

\begin{credits}
\subsubsection{\ackname} We acknowledge the CFIS Mobility Program for the partial funding of this research work, particularly Fundació Privada Mir-Puig, CFIS partners, and donors of the crowdfunding program. This publication has emanated from research conducted with the financial support of Research Ireland under Grant number 12/RC/2289\_P2.

\end{credits}

\newpage

\printbibliography[title={References}]

\end{document}